\documentclass{article}

\usepackage{PRIMEarxiv}
\usepackage{algorithm}
\usepackage{algorithmic}

\usepackage[utf8]{inputenc} 
\usepackage[T1]{fontenc}    
\usepackage{hyperref}       
\usepackage{url}            
\usepackage{booktabs}       
\usepackage{amsfonts}       
\usepackage{nicefrac}       
\usepackage{microtype}      
\usepackage{lipsum}
\usepackage{fancyhdr}       
\usepackage{graphicx}       
\graphicspath{{media/}}     
\usepackage{amssymb}
\usepackage{amsmath}
\usepackage{subfig}
\usepackage{caption}
\usepackage{appendix}
\usepackage{hyperref}
\usepackage[bottom]{footmisc}

\title{FRIDA - Generative feature replay for incremental domain adaptation}

\author{
  Sayan Rakshit\\
  \texttt{sayan1by2@gmail.com}\\
  Indian Institute of Technology Bombay, \\Mumbai, 400076, India\\
  \And
  Anwesh Mohanty\\
  \texttt{anweshm136@gmail.com}\\
  Indian Institute of Technology Bombay, \\Mumbai, 400076, India\\
  \And
  Ruchika Chavhan\\
  \texttt{chavhanruchika2801@gmail.com }\\
  Indian Institute of Technology Bombay, \\Mumbai, 400076, India\\
  \And
  Biplab Banerjee\\
  \texttt{getbiplab@gmail.com}\\
  Indian Institute of Technology Bombay, \\Mumbai, 400076, India\\
  \And 
  Gemma Roig\\
  \texttt{roig@cs.uni-frankfurt.de}\\
  Goethe University Frankfurt, \\Frankfurt, 60325, Germany\\
  \And
  Subhasis Chaudhuri\\
  \texttt{sc@ee.iitb.ac.in}\\
  Indian Institute of Technology Bombay, \\Mumbai, 400076, India\\
}

\begin{document}
\maketitle
\begin{abstract}
We tackle the novel problem of incremental unsupervised domain adaptation (IDA) in this paper. We assume that a labeled source domain and different unlabeled target domains are incrementally observed with the constraint that data corresponding to the current domain is only available at a time. The goal is to preserve the accuracies for all the past domains while generalizing well for the current domain. The IDA setup suffers due to the abrupt differences among the domains and the unavailability of past data including the source domain. Inspired by the notion of generative feature replay, we propose a novel framework called Feature Replay based Incremental Domain Adaptation (FRIDA) which leverages a new incremental generative adversarial network (GAN) called domain-generic auxiliary classification GAN (DGAC-GAN) for producing domain-specific feature representations seamlessly. For domain alignment, we propose a simple extension of the popular domain adversarial neural network (DANN) called DANN-IB which encourages discriminative domain-invariant and task-relevant feature learning. Experimental results on Office-Home, Office-CalTech, and DomainNet datasets confirm that FRIDA maintains superior stability-plasticity trade-off than the literature. \let\thefootnote\relax\footnotetext{Accepted at "Computer Vision and Image Understanding",7th January 2022.}
\end{abstract}


\section{Introduction}

\begin{figure*}[h]
    \centering
    \includegraphics[width=0.95\linewidth, height=5cm]{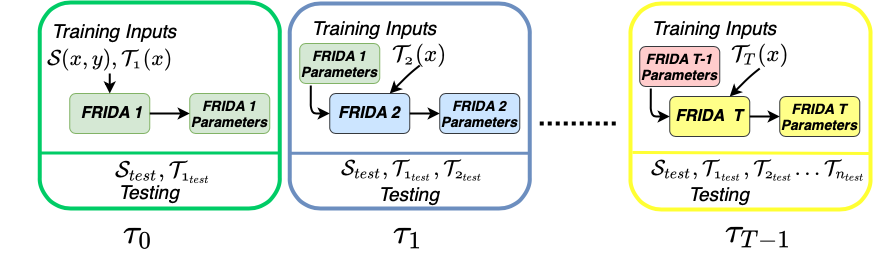}
    \caption{The figure depicts the problem of IDA and how FRIDA works in this setup. The labeled source domain $\mathcal{S}$ arrives as $\tau_0$ and the unlabeled target domains $\mathcal{T}_{\tau}$ follow in subsequent time steps. The FRIDA model at a given $\tau$ is initialized from the most recent previous model of $\tau-1$ and in this way the past information propagates temporally and avoids forgetting. At any $\tau$, testing can be done on the domains seen so far and the domain identity is unavailable for the test samples.}
    \label{fig:pic1}
\end{figure*}
The deep learning techniques have manifested impressive performances for a wide range of visual inference tasks (\cite{1,2}). Yet, there persist two vital issues with deep convolutional neural networks (CNN).
First, these models often find it onerous to generalize confidently when exposed to a new environment if there exists distributions-gap (\cite{3}). The domain adaptation (DA)  (\cite{4,5}) techniques come to the rescue in such a situation. 
 Another problem surfaces when task-specific data are obtained in an online fashion. Although it is plausible for human beings to acquire new knowledge without impeding the existing skills, machine learning techniques are prone to overwrite the parameters learned based on the previous data once they experience new information. Generally, access to data related to old tasks may be restricted due to several reasons. This hinders the possibility to maintain the performance on the old tasks persistently which is known as the catastrophic forgetting (\cite{9}) of incremental learning. Nevertheless, several measures have been introduced to judiciously combat the forgetting issue in neural networks (\cite{7, 12, 13, 15}).
 

In this paper, we bring in a new incremental learning paradigm for the unsupervised DA problem (IDA) to carry out the task of image classification. In unsupervised DA, there exists a labeled source domain $\mathcal{S}$ and an unlabeled target domain $\mathcal{T}$ in pair with $P(\mathcal{S}) \neq P(\mathcal{T})$. In contrast, the proposed non-stationary IDA setup considers that the domains are available progressively. While the labeled $\mathcal{S}$ arrives at $\tau=0$, different unlabeled target domains $\{\mathcal{T}_{\tau}\}_{\tau=1}^T$ appear at different time-stamps. Data corresponding to the current domain are only available at a given time which means $\mathcal{S}$ cannot be directly accessed at $\tau \geq 1$. Further, we do not anticipate any constraint on the domain-shifts.
The proposed setup applies to the real-world classification problems on memory-constrained systems where data from the same set of categories can come temporally from varied sources and there is a lack of available annotations. Some possible application areas include robot vision, medical imaging, remote sensing. For example, a robot with limited memory capacity may be trained with synthetic data and is subsequently deployed into real environments where the robot has to navigate through different rooms. Here, the labeled synthetic domain acts as $\mathcal{S}$ where the different room in the real space signify the $\mathcal{T}_{\tau}$s. The goal is that the robot should be able to memorize all the real environments throughout. 
Although there exists barely a few IDA techniques in the literature (\cite{11,44,45}),
they cannot handle our IDA setup since either they require $\mathcal{S}$ to be always available or assume some homogeneity across all the source and target domains (like insignificant domain shift).

Despite the unavailability of data from past domains, we argue for trickily solving IDA by adapting every new target domain using an unsupervised DA technique. For that, we have to consider some representative labeled samples from the previous domains
to serve as the replay memory in the form of an auxiliary source domain. Although data can be stored for the past domains, obtaining the labels is non-trivial given that the target domains are devoid of annotations. While we can store the original training data for $\mathcal{S}$, pseudo-labeling can be carried out to approximate the labels for the target domains.
However, this starkly contrasts with the natural learning mechanisms of the
human brain, which does not consider the retrieval of raw information similar to the originally exposed impressions (\cite{53}). Hence, storing domain-specific real samples is not encouraged.

To this end, the notion of non-rehearsal based generative replay to avoid the forgetting problem has lately gained popularity for class incremental learning (CIL) (\cite{18}). In this case, a deep generative model is trained to produce synthetic data for the old classes continuously. Joint training is henceforth carried out given these generated data along with the labeled training data of the novel classes. Some initial works (\cite{28}) focus on generating images for replay but training GAN for image generation in an incremental way is extremely hard. In contrast, some recent endeavors (\cite{19, 57}) switch to the notion of feature replay where a pre-trained feature extractor is used to generate visual representations from the images. This has gained popularity since handling GAN for feature vectors in a low-dimensional manifold is tractable than high-dimensional images. This persuades us to opt for the notion of feature replay to solve IDA but we would like to point out that such CIL approaches cannot be directly applied here given that IDA is different and more challenging than CIL. IDA is a label-deficit setup working with multiple diverse domains while CIL is a single domain supervised model.
Hence, we advocate the need to design a new generative feature replay scheme for IDA that should be able to output class-specific embeddings together with the pseudo class labels for different domains.


Motivated by the above arguments, we propose a novel generative modeling based IDA framework abbreviated as FRIDA (Feature Replay for Incremental Domain Adaptation) (Fig. \ref{fig:pic1}) which is the first of its kind. We use a common feature extractor for all the domains. As per the design, FRIDA consists of two novel modules for feature generation from the past domains and adapting the new domains and both the stages repeat alternately.
A domain generic AC-GAN (DGAC-GAN) is proposed as the replay memory which is incrementally trained to synthesize features for the past domains at every $\tau \geq 1$. DGAC-GAN has a novel architecture and it can be considered to be an extension of the standard AC-GAN model (\cite{58}) for multi-domain and multi-class data. For learning the joint distribution for multiple domains and classes, we propose to dually condition DGAC-GAN on the novel notion of domain information together with  class information whereas AC-GAN is only class-conditioned. Additional regularization is also imposed in DGAC-GAN to ensure a high correlation between the real and generated samples.

The motivation behind using DGAC-GAN generated samples instead of storing the original samples of the previous domains can be described in two points, (i) Storing of previous samples increase the data storing capacity (ii) At any given time-stamp $\tau$ the data may be unavailable due to data privacy issues or distortion of original data of previous domains or any other reason. For example, client1 has an annotated dataset of domain $ D_0$ at a time-stamp $\tau = 0$, client2 has an unannotated dataset from a domain $ D_1$ at time-stamp $\tau = 1$. But due to the reasons at (ii) $D_0$ is not available to the client2, then DGAC-GAN can generate the synthetic data of domain $D_0$ for performing adaptation with $D_1$. Again if at time-stamp $\tau=2$ client3 arrives with a domain $D_2$ then DGAC-GAN will produce the synthetic data of domains $D_0$ and $D_1$ to perform adaptation among $D_0$, $D_1$ and $D_2$. The storing of previous domain data violates the motivation of our IDA. The success of FRIDA mainly depends on the correct data generation ability of DGAC-GAN in the absence of the original data.

An unsupervised DA module called Information Bottleneck (IB) DANN (DANN-IB) is proposed to perform feature adaptation and classification given the synthetic features of the previous domains and the available data for the current domain.  For DA, we choose to deal with the DANN framework (\cite{20}) given its simplicity and robustness and inculcate two improvements that can be helpful for IDA.
 First, the IB principle (\cite{43}) is introduced in the DANN loss to encourage the latent space to be truly domain agnostic. Besides, we propose to utilize a multi-class discriminator in place of the standard binary discriminator of DANN to cope up with the multi-modal nature of the domains. A thresholding based pseudo-labeling is carried out based on the outputs of the DA stage to obtain confident domain samples to update DGAC-GAN. Our novelties are three-fold:

\noindent i) We tackle a novel IDA problem setup where  a single labeled source and multiple unlabeled target domains are dealt with considering the unavailability of information from all the past domains at every time and propose a framework called FRIDA.
ii) We rely on the notion of generative feature replay for this purpose and introduce a novel replay memory module called DGAC-GAN to output multi-domain features and a DA solver module called DANN-IB for adaptation and classification. Both are lightweight models which makes FRIDA easy to deploy in general.
iii)  We demonstrate the efficacy of FRIDA for three benchmark and large-scale datasets and rigorously ablate the model.

\section{Related works}

\noindent \textbf{Incremental learning}: Incremental learning (\cite{7}) is the process to continually train a classifier for sequentially available data for novel tasks by taking care of the catastrophic forgetting problem. The problem of class incremental learning is by far the most studied setup in this aspect.  
There exist three different types of IL approaches in the literature, regularization based, dynamic model architecture based, and exemplar replay based, respectively. While the regularization based methods (\cite {12, 22, 25}) penalize any change to the parameters important for the previous tasks, the dynamic modeling based approaches (\cite{23, 24}) increase the model capacity to accommodate new tasks.
The replay based methods either utilize real samples (\cite{26, 17}) (rehearsal) or synthetic samples (non-rehearsal) fabricated by some generative models (\cite{19}) corresponding to the previous tasks in order to prevent forgetting. 
Different versions of the conditional GAN architecture have been appraised to be the automatic choice for this purpose (\cite{19, 28, 29}).

\noindent \textbf{Incremental domain adaptation}: The computer vision literature is rich in sophisticated DA techniques for problems like object recognition, semantic segmentation, activity recognition, person re-identification (\cite{30}). Handling the discrepancy between the training and test data distributions is a long lasting problem in machine learning. In this regard, DA models are able to align the data distributions either implicitly through adversarial approaches or explicitly using standard divergence measures (\cite{31, 32, 33, 20, 38}).

The notion of IDA is comparatively less studied in the literature and is mainly designed for the scenario when an unsupervised DA model trained for the domain pair $(\mathcal{S}, \mathcal{T}_1)$ is to be updated for the pair $(\mathcal{S}, \mathcal{T}_2)$ without forgetting the previously learned mapping. The few existing works (\cite{40, 44}) in this line require prior information regarding the domains or their inter-relationships in terms of graph based structures. Besides, some of the techniques (\cite{45, 11}) are designed for continuously changing environments where the domain-shift is assumed to be marginal. 
Methods like (\cite{45, 11, 41}) utilize the notion of memory replay to control the forgetting by incorporating additional distillation loss functions or regularizers. However, these methods are not tailored to deal with the data-free IDA formulation introduced here given their over-reliance on the original source-domain samples.

Both the DANN-IB and DGAC-GAN models are different than the existing literature. For example, there exists a number of GAN models in the literature for multi-domain image to image translation (\cite{60,61}) or class incremental training (\cite{57}) but none can be utilized to handle the IDA setup. Basically, what makes DGAC-GAN different from them are: i) it can deal with multiple domains containing several classes continually as opposed to \cite{57} which can only perform class-specific feature generation for a given domain, ii) Although our non-stationary DGAC-GAN uses multiple conditioning to synthesize domain and class specific features, it is different from the translation models of \cite{60,61} which require paired cross-domain data to carry out the image translation task and are stationary in nature.
On the other hand, the adversarial training strategy of DANN has been historically extended for generating a discriminative shared feature space like in \cite{59,62}. However, \cite{59} is a heavy model with separate binary discriminators designed for each of the classes plus the training of \cite{59} is directed by the notion of pseudo-labeling. DANN-IB is a simple model with a $\mathcal{C}+1$-class discriminator and a single adversarial loss. What is more, DANN-IB integrates the notion of information bottleneck principle to eliminate the effects of domain-dependent artifacts which may jeopardise the alignment process, as opposed to other DA approaches.

\section{Proposed methodology}

\noindent \textbf{Preliminaries}: We assume the presence of $T+1$ visual domains $( \mathcal{T}_0, \mathcal{T}_1, \cdots, \mathcal{T}_T )$ where $\mathcal{T}_0$, also denoted as $\mathcal{S}$, represents the source domain while $\{\mathcal{T}_{\tau  \geq 1}\}$ defines the sequence of the target domains. In our setting, $\mathcal{S}$ is equipped with labeled training samples $\mathcal{D}_0 = \{\textbf{x}^i_0, y^i_0\}_{i=1}^{|\mathcal{D}_0|}$ whereas we have access to unlabeled samples for the target domains: $\mathcal{D}_{\tau} = \{\textbf{x}_{\tau}^j\}_{j=1}^{|\mathcal{D}_{\tau}|}$ for $\tau \geq 1$.  A given $\textbf{x}_{\tau} \in \mathbb{R}^d$ denotes the output of a pre-trained CNN which projects the images onto a $d$-dimensional feature space.
Ours is a closed-set setup where all the domains have samples from a fixed set of $\mathcal{C}$ semantic categories.
Furthermore, the different domains follow non-identical data distributions: $P(\mathcal{T}_i) \neq P(\mathcal{T}_j)$, $0 \leq i \neq j \leq T $ and we assume no prior knowledge regarding the domains.
Only $\mathcal{D}_{\tau}$ is considered to be available at $\tau$. Given that, we aim to design an adaptation cum classification framework that should be able to continuously adapt the new target domains while ensuring that the performance on the previous domains remains unhindered. Some of the important variables are depicted in Table \ref{tab:not_label}.

\subsection{Overview of the FRIDA framework}

\begin{figure*}[h]
    \centering
    \includegraphics[width=0.9\linewidth, height=5cm]{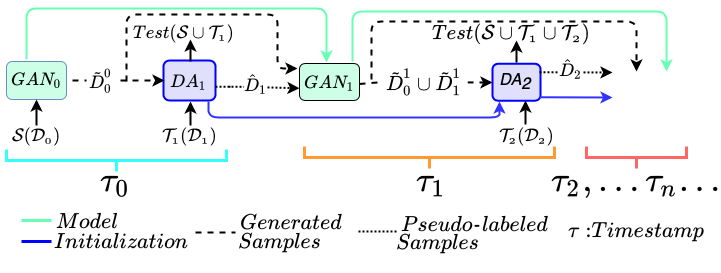}
    \caption{ The working of FRIDA for two time steps where the DGAC-GAN and the DANN-IB are interchangeably utilized. Details can be found in Section 3.1.}
    \label{fig:model}
\end{figure*}

While adapting a new target domain, FRIDA ensures the generation of synthetic features corresponding to the past source and target domains using DGAC-GAN and these samples act as the auxiliary source domain for the unsupervised DANN-IB. Both the models can be trained incrementally using the notion of joint training for accommodating new domains while preserving the past knowledge. With an abuse of notation, we denote the instances of DGAC-GAN and DANN-IB at time $\tau$ by $GAN_{\tau}$ and $DA_{\tau}$, respectively. 
In the following, we discuss the FRIDA workflow at $\tau=0,1$ and for $\tau \geq 2$ (also mentioned in Algorithm \ref{alg:the_alg}). 

A DGAC-GAN is first trained on $\mathcal{D}_0$ ($GAN_0$). At $\tau=1$, we have access to the unlabeled $\mathcal{D}_1$ from $\mathcal{T}_1$ but samples from $\mathcal{D}_0$ can no longer be obtained. At this point, 
$GAN_0$ is utilized to generate class-wise synthetic samples of $\mathcal{S}$ denoted by $\tilde{\mathcal{D}}_0^0$.
The DANN-IB module at $\tau=1$ ($DA_1$) is adversrially trained to align $\tilde{\mathcal{D}}_0^0$ and $\mathcal{D}_1$ and a $\mathcal{C}$-class classifier on $\tilde{\mathcal{D}}_0^0$ is modeled simultaneously. $DA_1$ can now handle test samples from $\mathcal{S} \cup \mathcal{T}_1$. Our focus now is to obtain $GAN_1$ that can subsequently produce synthetic samples both for $\mathcal{S}$ and $\mathcal{T}_1$ at $\tau=2$. However, $\mathcal{D}_1$ is unlabeled while DGAC-GAN requires label information during training. As a remedy, we propose a pseudo-labeling stage to obtain a subset of highly confident samples denoted by $\hat{\mathcal{D}}_1$ from $\mathcal{D}_1$ by thresholding the classifier's outputs of $DA_1$.
$GAN_1$ is henceforth trained by considering $\tilde{\mathcal{D}}_0^0 \cup \hat{\mathcal{D}}_1$ as real (Fig. \ref{fig:model}). 

For $\tau \geq 2$, $GAN_{\tau-1}$ is used to generate synthetic samples along with the pseudo labels denoted by $\tilde{\mathcal{D}}_0^{\tau-1} \cup \tilde{\mathcal{D}}_1^{\tau-1} \cup \cdots \cup \tilde{\mathcal{D}}_{\tau-1}^{\tau-1}$ for $\mathcal{S}$ and the $\tau -1$ target domains. Now for adapting the generated data with $\mathcal{D}_{\tau}$, $DA_{\tau}$ considers $\tilde{\mathcal{D}}_0^{\tau-1} \cup \tilde{\mathcal{D}}_1^{\tau-1} \cup \cdots \cup \tilde{\mathcal{D}}_{\tau-1}^{\tau-1}$ to constitute an auxiliary source domain $\mathcal{S}_{\tau}$ while $\mathcal{D}_{\tau}$ acts as the target $\mathcal{T}_{\tau}$. After training, we obtain $\hat{\mathcal{D}}_{\tau}$ using pseudo-labeling and subsequently train $GAN_{\tau}$ with  $\tilde{\mathcal{D}}_0^{\tau-1} \cup \tilde{\mathcal{D}}_1^{\tau-1} \cup \cdots \cup \tilde{\mathcal{D}}_{\tau-1}^{\tau-1} \cup \hat{\mathcal{D}}_{\tau}$. 
For modelling $GAN_{\tau}$ or $DA_{\tau}$, we initialize the parameters of the models from the snapshots of $GAN_{\tau-1}$ and $DA_{\tau-1}$. Details of DGAC-GAN and DANN-IB are mentioned below.

\begin{table}[]
    \centering
    \begin{tabular}{|c|c|}
    \hline
    
         $\mathcal{S(D}_{0})$,
         $\mathcal{T(D}_{\tau  \geq 1}) $&  Source Domain, Target Domains  \\
         \hline
         
         $\tilde{\mathcal{D}}_0^0$& Gan generated Source domain at $\tau = 0$ \\
         \hline
         
         $\tilde{\mathcal{D}}_T^\tau$& Gan generated samples of different \\&domains($\tau \geq 0$) at different timestamp $T \geq 0  $ \\
         \hline
         
         $\hat{\mathcal{D}}_{\tau\geq 1}$& Pseudo Labeled from dann w.r.t each target \\&domains ($\tau\geq 1$) at that particular timestamp  \\
         \hline
         [$\textbf{z},y,\tau$]& [$\textbf{z} \in \mathcal{N}(0,\mathbb{I})$ (noise), class labels, domain labels] \\
         \hline
    \end{tabular}
    \caption{List of important notation}
    \label{tab:not_label}
\end{table}


The FRIDA is composed of the knowledge of DGAC-GAN and DANN-IB, where DGAC-GAN is dedicated to generating the samples of previous domains from noise, and DANN-IB is responsible for performing the adaptation between newly arrived domains with the generated samples of the previous domains. Our DGAC-GAN is capable of generating the samples according to the class labels as well as domain labels which differs from the AC-GAN, where it only be able to generate samples according to their class labels. Since in IDA setup, we work with multiple domains, so we need to generate samples according to their domain labels along with the class labels as well.

The DANN-IB is an extension of DANN. Instead of considering a binary domain classifier similar to DANN which fails to preserve the discriminative nature of classes in the source domain, our DANN-IB aligns the domains while preserving the discriminative nature of source categories by adversarially discriminating among the source classes $(1, 2, ... , C)$ and the target domain $(C+1)$.

\subsection{DGAC-GAN}

\begin{figure*}[ht]
    \centering
    \begin{minipage}[b]{0.47\linewidth}
    \fbox{\includegraphics[width=0.95\linewidth, height=4cm]{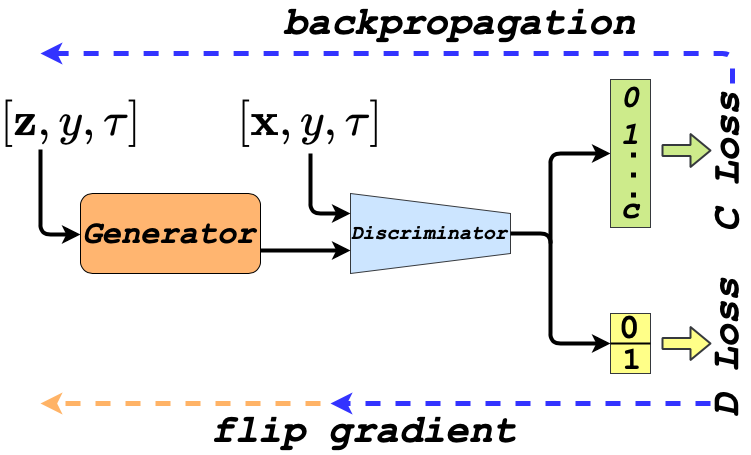}}
    \caption{ The DGAC-GAN model with generator, discriminator sub-networks, and the two classifier heads.}
    \label{fig:model_gan}
    \end{minipage}
 \quad
 \begin{minipage}[b]{0.47\linewidth}
    \fbox{\includegraphics[width=0.97\linewidth, height=3.97cm]{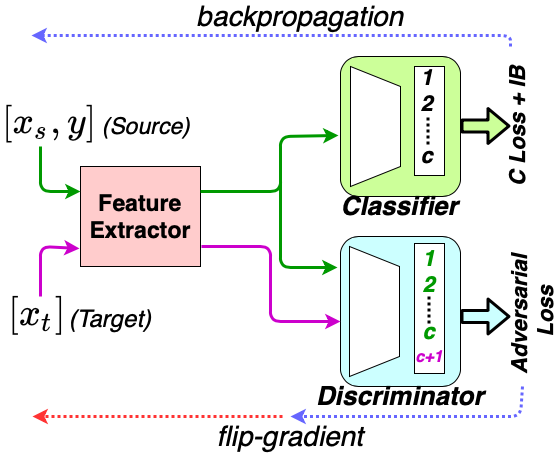}}
    \caption{DANN-IB model with the feature encoder and the two classifier heads.  $~~~~~~~~~~~~~~~~~~~~~~~~~ $                                                  }
    \label{fig:model_dann}
    \end{minipage}
\end{figure*}

As mentioned, formalizing a GAN as the replay memory for IDA setup is challenging mainly because of the discrepancies among the domains. A naive possibility would be to use the dynamically expandable GAN models where new domain-specific generators and discriminators are to be added continuously. However, such a model may have scalability issue for a large number of domains. Instead, we propose DGAC-GAN (Fig. \ref{fig:model_gan}) which maintains a static architecture over the episodes and can be easily trained incrementally.

 By design, DGAC-GAN consists of a feature generator $\mathcal{G}^{gan}(;,\theta_{\mathcal{G}}^{gan})$, a binary real/fake discriminator $f_D^{gan}(;, \theta_D^{gan})$ and a auxiliary multi-class classifier $f_{C}^{gan}(;, \theta_{C}^{gan})$. 
 Both $f^{gan}_D$ and $f^{gan}_C$ share a common discriminator sub-network $f_{sh}^{gan}(;,\theta_{sh}^{gan})$ from which two different classification heads are enacted. We note that the $\theta^{gan}$s represent the model parameters.
 In order to handle multiple domains, we introduce the notion of domain identifier where a given $\mathcal{T}_{\tau}$ can be recognized by an encoding (e.g. one-hot or binary) of the respective time-stamp $\tau$. For ease of understanding, we represent the encoded domain identifier vector by $\tau$ itself.

 Given noise samples $\textbf{z} \in \mathcal{N}(0, \mathbb{I})$, the input to the generator network is the concatenated vector $[\textbf{z},y, \tau]$. It is worth mentioning that the latent vector $\textbf{z}$ is completely agnostic to the domain and the category, and the same $\textbf{z}$ can be used to generate classwise features of different domains just by using different $y$ and $\tau$.
On the other hand, the inputs to $f_{sh}^{gan}$ are the real $[\textbf{x},\tau]$($d$-dimensional feature vector) and the generated $[\mathcal{G}^{gan}([\textbf{z},y,\tau]),\tau]$ examples both conditioned on $\tau$. 
Although the traditional AC-GAN model does not mention any conditioning while feeding data to the discriminator, however, we choose to condition $f_{sh}^{gan}$ on $\tau$ in DGAC-GAN so that the discriminator is informed regarding the domain information barring the need to dynamically expand the network for new domains.
The objective function of DGAC-GAN comprises of two parts: the log-likelihood of the i) correct data source which signifies whether a given sample is obtained from the original distribution or generated by $\mathcal{G}^{gan}$: $ real (r) / fake (f)$ ($\mathcal{L}_s^{gan}$) and, the ii) correct class label ($\mathcal{L}_c^{gan}$) given the $\mathcal{C}$ categories, respectively, with $\textbf{x}' = \mathcal{G}^{gan}(\textbf{z},y,\tau)$.

\begin{equation}
\begin{array}{lll}
    \mathcal{L}_s^{gan} = \mathbb{E} [\log P(f_D^{gan}(\textbf{x}) = r | \textbf{x}) + \log P(f_D^{gan}(\textbf{x}') = f | \textbf{z})]
\\ \\
    \mathcal{L}_c^{gan} = \mathbb{E} [\log P(f_{C}^{gan}(\textbf{x}) = y | \textbf{x}) + \log P(f_{C}^{gan}(\textbf{x}') = y | \textbf{z})] 
     \end{array}
     \label{eq:1}
\end{equation}

In order to guarantee an improved overlapping between the original and generated samples classwise, an $\ell_2$-norm based regularizer is also considered as follows,

\begin{equation}
    \centering
    \mathcal{R}^{gan} = \mathbb{E}_{x \sim (\tau, y)} ||\mathcal{G}^{gan}([\textbf{z},y,\tau]) - \textbf{x}||_2^2
\end{equation}

We follow the usual min-max training strategy for optimizing DGAC-GAN where $\{\theta_D^{gan}, \theta_{sh}^{gan}, \theta_{C}^{gan}\}$ are trained by maximizing $\mathcal{L}_s^{gan} + \mathcal{L}_c^{gan}$ 
and $\theta_{\mathcal{G}}^{gan}$ is trained by maximizing $ \mathcal{L}_c^{gan} - \mathcal{L}_s^{gan} - \mathcal{R}^{gan}$. 
As previously mentioned, we initialize $GAN_{\tau}$ from a snapshot of $GAN_{\tau-1}$. Since $GAN_{\tau-1}$ is already trained for $\mathcal{S} \cup \mathcal{T}_1 \cup \cdots \cup \mathcal{T}_{\tau-1}$, it becomes comparatively less demanding for $GAN_{\tau}$ to incorporate the additional knowledge of $\mathcal{T}_{\tau}$ through $\hat{\mathcal{D}}_{\tau}$ over the existing knowledge of $GAN_{\tau-1}$.

\subsection{DANN-IB and pseudo-labeling}

At time $\tau$, the goal of the domain adaptation stage is to project the labeled synthetic samples of the previous domains as generated by $GAN_{\tau-1}$:
  $\tilde{\mathcal{D}}_0^{\tau-1} \cup \tilde{\mathcal{D}}_1^{\tau-1} \cup \cdots \cup \tilde{\mathcal{D}}_{\tau-1}^{\tau-1}$ and the unlabeled samples of the current domain $\mathcal{D}_{\tau}$ in a common feature space such that the classifier trained on $\tilde{\mathcal{D}}_0^{\tau-1} \cup \tilde{\mathcal{D}}_1^{\tau-1} \cup \cdots \cup \tilde{\mathcal{D}}_{\tau-1}^{\tau-1}$ can be utilized to classify samples from all the $\tau$ domains. The DA module considered in FRIDA is designed for the single-source single-target setup and to comply with the same, 
  $\tilde{\mathcal{D}}_0^{\tau-1} \cup \tilde{\mathcal{D}}_1^{\tau-1} \cup \cdots \cup \tilde{\mathcal{D}}_{\tau-1}^{\tau-1}$ is presumed to constitute an auxiliary source domain $\mathcal{S}_{\tau}$ with labeled samples $(\mathcal{X}_{\mathcal{S}}^{\tau}, \mathcal{Y}_{\mathcal{S}}^{\tau})$ while $\mathcal{D}_{\tau}$ denotes the target domain $\mathcal{T}_{\tau}$ with unlabeled samples $\mathcal{X}_{\mathcal{T}}^{\tau}$.  
  We first discuss the traditional DANN model and subsequently DANN-IB is elaborated.
Given $\mathcal{S}_{\tau}$ and $\mathcal{T}_{\tau}$, DANN comprises of a feature extractor $\mathcal{G}^{dann} (;, \theta_{\mathcal{G}}^{dann})$, a domain classifier $f_{D}^{dann}(;, \theta_{D}^{dann})$, and a source domain specific multi-class classifier $f_{C}^{dann}(;, \theta_C^{dann})$ with the learnable parameters $\{\theta^{dann}\}$. To address the domain-shift problem, DANN minimizes the following objective function\ref{Eq3}, with respect to  $\{\theta_D^{dann}, \theta_C^{dann}, \theta_{\mathcal{G}}^{dann}\}$

\begin{equation}
    {\min}\{\mathcal{L}_c^{dann}(f_{C}^{dann}(\mathcal{G}^{dann}(\mathcal{X}_{\mathcal{S}}^{\tau})), \mathcal{Y}_{\mathcal{S}}^{\tau}) - \lambda (\mathcal{L}_d^{dann}(f_{D}^{dann}(\mathcal{G}^{dann}(\mathcal{X}_{\mathcal{S}}^{\tau})), 0) + \mathcal{L}_d^{dann}(f_{D}^{dann}(\mathcal{G}_{D}^{dann}(\mathcal{X}_{\mathcal{T}}^{\tau})), 1))\}
\label{Eq3}
\end{equation}

\noindent where $\mathcal{L}_c^{dann}$ and $\mathcal{L}_d^{dann}$ denote the multi-class classification loss for samples in $\mathcal{S}_{\tau}$ and the binary domain classification loss for ($\mathcal{X}^{\tau}_{\mathcal{S}},\mathcal{X}^{\tau}_{\mathcal{T}}$), respectively. $\lambda$ is the trade-off parameter and is updated in an adaptive fashion as per \cite{20}.
The ground-truth domain labels for $\mathcal{S}_{\tau}$ and $\mathcal{T}_{\tau}$ are set to $0$ and $1$ for $\mathcal{L}_d^{dann}$.
Intuitively, $f_{D}^{dann}$ is trained to distinguish the latent representations of $\mathcal{S}_{\tau}$ from those of $\mathcal{T}_{\tau}$, while $\mathcal{G}^{dann}$ is jointly trained to confuse $f_{D}^{dann}$ by maximizing $\mathcal{L}_d^{dann}$ while minimizing $\mathcal{L}_c^{dann}$ to avoid any trivial solution. 
 However, merely matching the marginal domain distributions in the latent space is not appreciative enough for ensuring the discriminative knowledge to be transferred from $\mathcal{S}_{\tau}$ to $\mathcal{T}_{\tau}$. It is highly likely that $\mathcal{G}^{dann}$ is misled by the domain invariant yet task irrelevant factors and fails to capture the semantic shared information. The problem is very relevant in our IDA setup where $\mathcal{S}_{\tau}$ at every $\tau \geq 2$ is considered to contain samples from multiple diverse domain distributions having significantly different domain-centric artifacts.
From another point of view, $f_{D}^{dann}$ in DANN performs global distributions matching between $\mathcal{S}_{\tau}$ and $\mathcal{T}_{\tau}$ since it is designed as a binary classifier. As a result, this overlooks the multi-modal nature of the domain samples and may promote misclassification for fine-grained categories. 

In order to mitigate the aforesaid drawbacks of DANN, we introduce an improved DANN model where i) $f_{D}^{dann}$ is designed to be a $\mathcal{C}+1$-class classifier instead of a binary classifier where the samples from $\mathcal{T}_{\tau}$ are assigned the class label $\mathcal{C}+1$ and the samples from $\mathcal{S}_{\tau}$ use the original $\mathcal{C}$ labels. The adversarial training now assigns one of the $\mathcal{C}$ labels to each of the target samples, thus assisting in a fine-grained domain alignment between $\mathcal{S}_{\tau}$ and $\mathcal{T}_{\tau}$ by minimizing the entropy of the predictions. ii) the information bottleneck (IB) (\cite{43}) regularizer is incorporated in the latent representations which enforces $\mathcal{G}^{dann}$ to ignore the task irrelevant features and focuses only on preserving the minimal sufficient statistics of the input data with respect to $\mathcal{S}_{\tau}$ and $\mathcal{T}_{\tau}$ (Figure \ref{fig:model_dann}). We consider the variational information bottleneck (VIB) (\cite{42}) principle in this respect and re-design $\mathcal{G}^{dann}$ as a stochastic feature extractor which maps a certain input sample into a Gaussian distributed latent space. We seek to minimize the mutual information between the original features and the latent representations. However given the intractable nature of the problem, we instead optimize the variational upper bound.
The following regularizer $\mathcal{R}_{IB}$ constrains the latent representations to follow the standard normal distribution $\mathcal{Q} = \mathcal{N}(0, \mathbb{I})$ as per \cite{42},

\begin{equation}\label{eqn:four}
    \mathcal{R}_{IB} =  D_{KL}(Q(\mathcal{G}^{dann}(\mathcal{X}_{\mathcal{S}}^{\tau}))|| \mathcal{Q}) + D_{KL}(Q(\mathcal{G}^{dann}(\mathcal{X}^{\tau}_{\mathcal{T}}))|| \mathcal{Q})
\end{equation}

\noindent In Eq. (\ref{eqn:four}), $D_{KL}$ is the Kullback-Leibler divergence between the respective distributions. The overall loss function for DANN-IB is in Eq. (\ref{eqn:five}) minimize with respect to $\{\theta_{\mathcal{G}}^{dann},\theta_{\mathcal{C}}^{dann}, \theta_{\mathcal{D}}^{dann}\}$.

\begin{equation}
\min\{ \ \mathcal{L}_c^{dann}(f_{C}^{dann}(\mathcal{G}^{dann}(\mathcal{X}_{\mathcal{S}}^{\tau})), \mathcal{Y}_{\mathcal{S}}^{\tau}) - \lambda (\mathcal{L}_d^{dann}(f_{D}^{dann}(\mathcal{G}^{dann}(\mathcal{X}_{\mathcal{S}}^{\tau})), Y_{\mathcal{S}}^{\tau}) +  \mathcal{L}_d^{dann}(f_{D}^{dann}(\mathcal{G}^{dann}(\mathcal{X}_{\mathcal{T}}^{\tau})), \mathcal{C}+1)) +  \mathcal{R}_{IB}\}
\label{eqn:five}
\end{equation}

The next step is to obtain pseudo-labeled samples corresponding to $\mathcal{T}_{\tau}$ which can further be used to obtain $GAN_{\tau}$ from $GAN_{\tau-1}$. Obtaining confident pseudo-labels is not straight-forward in the context of domain adaptation. One simple way is to identify the confident samples where the confidence is measured in terms of class membership values.
 Using this idea, we pass $\mathcal{D}_{\tau}$ through the trained DANN-IB network to record the class probabilities as provided by $f_{C}^{dann}$. We construct the subset of highly confident target samples  with pseudo labels $\hat{D}_{\tau}$ by selecting samples for each of the class labels where the class posterior probability exceeds a pre-defined threshold value $Th$. The analysis of the $\mathcal{H}\Delta \mathcal{H}$ for DANN-IB is provided in the Appendix 5.
\begin{algorithm}[H]
\caption{Working principle of FRIDA}
\begin{algorithmic}[1]
\label{alg:the_alg}
\renewcommand{\algorithmicrequire}{\textbf{Input:}}
\REQUIRE $\mathcal{D}_0$, $\mathcal{D}_1$,$\mathcal{D}_2$,$\cdots$,$\mathcal{D}_T$ and threshold $Th$
\renewcommand{\algorithmicrequire}{\textbf{Output:}}
\REQUIRE  The trained $GAN_{\tau}$ and $DA_{\tau}$ at each $\tau \in [1,T]$
\IF{$\tau=0$}
\STATE Train $GAN_0$ on $\mathcal{D}_0$ 
\ELSIF{$\tau = 1$}
\STATE Generate $\tilde{\mathcal{D}}_0^0$ using $GAN_0$
\STATE Train $DA_1$ with $\mathcal{S}_1$ as $\tilde{\mathcal{D}}_0^0$ and $\mathcal{T}_1$ as $\mathcal{D}_1$ (Eq. 5)

\STATE Apply pseudo-labeling with threshold $Th$ to obtain $\hat{\mathcal{D}}_1$ from $\mathcal{D}_1$
\STATE Obtain $GAN_1$ using  $\tilde{\mathcal{D}}_0^0 \cup \hat{\mathcal{D}}_1$
 as the real data
\ELSIF{$\tau \geq 2$}
\STATE Obtain $\tilde{\mathcal{D}}_0^{\tau-1} \cup \tilde{\mathcal{D}}_1^{\tau-1} \cup \cdots \cup \tilde{\mathcal{D}}_{\tau-1}^{\tau-1}$ using $GAN_{\tau-1}$
\STATE Train $DA_{\tau}$ with $\mathcal{S}_{\tau} = \tilde{\mathcal{D}}_0^{\tau-1} \cup \tilde{\mathcal{D}}_1^{\tau-1} \cup \cdots \cup \tilde{\mathcal{D}}_{\tau-1}^{\tau-1}$ and $\mathcal{D}_{\tau}$ as $\mathcal{T}_{\tau}$ (Eq. 5)
\STATE Obtain $\hat{\mathcal{D}}_{\tau}$ using pseudo-labeling with $Th$
\STATE Obtain $GAN_{\tau}$ by considering $\tilde{\mathcal{D}}_0^{\tau-1} \cup \tilde{\mathcal{D}}_1^{\tau-1} \cup \cdots \cup \tilde{\mathcal{D}}_{\tau-1}^{\tau-1} \cup \hat{\mathcal{D}}_{\tau}$ as the real data 
\ENDIF
\end{algorithmic}
\end{algorithm}

\begin{table*}
\centering
\scriptsize
\begin{tabular}{|c|c|c|c|c|c|c|c|}
\hline
Method&$\mathcal{T}_1$  & $\mathcal{T}_2$ & $\mathcal{T}_3$ & $\mathcal{T}_4$ & $\mathcal{T}_5$ & Target(AVG)& Source (AVG) \\
&A~~~~F & A~~~~F & A~~~~F & A~~~~F & A & A ~~~~~~~~~~~~~~~~~~~~F& A~~~~~~~~~~~~~~~~~~~~F\\

\hline DANN 1(baseline)&20.88~~-7.89 &14.00~~-4.96 &7.45~~-3.85 &1.12~~-0.12 &1.27 &8.94($\pm 0.30$)~~~~~ -4.21&39.14$(\pm 0.20)$~~-17.36\\
\hline
\hline DANN 2 (baseline) &33.01~~+0.05&31.63~~~0.00&18.51~~+0.10&2.09~~+0.40&10.86&19.22$^*(\pm 0.25)$~+0.14&74.28$^*$~~~~~~~~~~~~~~~~-\\

\hline DADA \cite{63}&-&-&-&-& & 20.7±0.4* ~~~~~~~ -&-\\
\hline
\hline IADA \cite{44} &28.58~~-0.74&24.41~~-0.49 &14.46~~-0.77 &1.47~~-0.40 &8.35 &15.45 $(\pm0.30)$~~~-0.60&74.71$^*$~~~~~~~~~~~~~~~~-\\

\hline CUA \cite{11} &20.92~~-7.39 &13.24~~-3.88 &7.32~~-2.53 &1.05~~-0.24 &1.81 &8.87$(\pm 0.40)$~~~~~ -3.51&39.38$(\pm 0.35)$~~-16.55\\
\hline
\hline EWC\cite{12}&20.33~~-7.93 &13.03~~-4.78&7.25~~-3.37&1.01~~-0.17&0.88 &8.50$(\pm 0.25)$~~~~~ -4.06 &36.80$(\pm 0.40)$~~-17.43  \\
\hline LWF\cite{25}&20.62~~-7.94 &13.26~~-4.52 &7.49~~-3.41 &1.25~~-0.17&1.15 &8.75$(\pm 0.35)$~~~~~ -4.01 &37.60$(\pm 0.50)$~~-17.34\\
 
\hline
\hline \textbf{Ours(FRIDA-DANN})&27.80~~-2.44&24.73~~-3.06 & 15.05~~-1.10& 1.67~~-0.08&8.80 &15.61$(\pm 0.40)$~~~~-1.67&63.17$(\pm 0.50)$~~-3.34\\
\hline \textbf{Ours(FRIDA-DANN-IB)}&\textbf{29.92~~-1.13} &\textbf{28.19~~-1.05} &\textbf{16.79~~-0.91} &\textbf{2.84~~+0.10} & \textbf{8.98}~~&\textbf{17.34$(\pm 0.25)$~~~~-0.75}&\textbf{64.71$(\pm0.30)$~~-2.38} \\
\hline
\end{tabular}
\caption{The performance comparison on the DomainNet dataset. We report the average performance (A) and forgetting (F) for each domain as well as the entire dataset. A negative average forgetting denotes that on an average, there is a performance drop for the domain. '-' signifies that the accuracy values are constant and * means it is not an incremental setup. (In \%)}\label{tab:1}
\end{table*}


\begin{table*}
\centering
\scriptsize
\begin{tabular}{|c|c|c|c|c|c|}
 \hline

\hline
Method& $\mathcal{T}_1$  & $\mathcal{T}_2$ & $\mathcal{T}_3$  &T(AVG)&S(AVG)\\
&A~~~~F&A~~~~F&A& A~~~~~~~~~~F & A~~~~~~~~~~F\\

\hline DANN 1(baseline)  &73.42~~-5.70 &45.30~~+0.23 &45.40 &54.71~~-2.73&79.18~~-10.39\\
\hline
\hline DANN 2(baseline)   &76.98~~+0.83&48.13~~+1.14 & 63.51& 62.87$^*$+0.99&89.48$^*$~~~~~- \\
\hline DADA\cite{63}  &- &- &- &59.5±0.2*~~~-&  \\
\hline
\hline IADA \cite{44} &75.60~~+1.31 &46.18~~~0.00 & 59.26&60.35~~+0.65&89.37$^*$~~~~~- \\
\hline CUA \cite{11} &76.10~~-1.35 &47.45~~+1.50 &55.42 &59.66~~+0.07&82.43~~-3.89\\
\hline
\hline EWC \cite{12} &73.22~~-6.25 &46.10~~+1.37 &47.33 &55.55~~-2.69&80.63~~-8.83  \\
\hline LWF\cite{25}& 72.20~~-5.33&44.47~~+1.45 &50.48 &55.72~~-1.94&79.84~~-10.29 \\
\hline
\hline {\textbf{Ours(FRIDA(DANN))}} &76.06~~-1.46&63.40~~+0.70&66.20 &68.55~~-1.08&84.56~~-1.57\\
\hline {\textbf{Ours(FRIDA(DANN-IB))}}  &\textbf{77.40~~-0.41}&\textbf{64.31~~+2.06}&\textbf{67.76}&\textbf{69.82~~+0.83}&\textbf{83.16~~-0.80} \\
\hline
\end{tabular}
\caption{The performance comparison using average accuracy (A) and forgetting (F) for Office-Home datasets. '-' signifies that the accuracy values are constant for all the episodes. * means it is not an incremental setup. (In \%)}
\label{tab:2}
\end{table*}


\begin{table*}
\centering
\scriptsize
\begin{tabular}{|c|c|c|c|c|c|}
 \hline

\hline
Method& $\mathcal{T}_1$  & $\mathcal{T}_2$ & $\mathcal{T}_3$  &T(AVG)&S(AVG)\\
&A~~~~F&A~~~~F&A& A~~~~~~~~~~F & A~~~~~~~~~~F\\

\hline DANN 1 (baseline)  &94.44~~~0.00 &82.03~~-4.49 &70.92&82.46~~-2.25&84.84~~-7.63 \\
\hline
\hline DANN 2 (baseline) &95.14~~-1.05 &83.14~~+11.23 &91.40 &89.89$^*$+5.09 &96.53$^*$~~~- \\
\hline DADA\cite{63}  &- &- &-& 92.0±0.4*~~ -&- \\
\hline
\hline IADA \cite{44} &95.14~~-1.04 &85.39~~-2.25 &87.83 &89.45~~-1.65 &96.87$^*$~~~-  \\
\hline CUA \cite{11} &95.13~~-1.04 &84.83~~+1.12 &80.71&86.89~~+0.04&88.89~~-4.34 \\
\hline
\hline EWC \cite{12} &92.36~~-3.13&84.83~~-1.12 &76.56 &84.58~~-2.13&82.41~~-6.08 \\
\hline LWF\cite{25}&95.84~~-1.05 &85.95~~-1.13 &82.49 &88.09~~-0.55&84.84~~-6.77 \\
\hline
\hline \textbf{Ours(FRIDA(DANN))} &95.20~~-1.75&95.71~~-2.10 &87.50 &92.80~~-1.93&94.78~~-0.32 \\
\hline \textbf{Ours(FRIDA(DANN-IB))} &\textbf{97.67~-1.03}&\textbf{99.07~-1.87}&\textbf{88.42} &\textbf{95.05~~-1.45}&\textbf{95.03~~+0.23}\\
\hline
\end{tabular}
\caption{The performance comparison using average accuracy (A) and forgetting (F) for Office-Caltech. '-' signifies that the accuracy values are constant for all the episodes. * means it is not an incremental setup. (In \%)}
\label{tab:3}
\end{table*}



\begin{figure*}[h]
    \centering
    \includegraphics[width=1\linewidth, height=5cm]{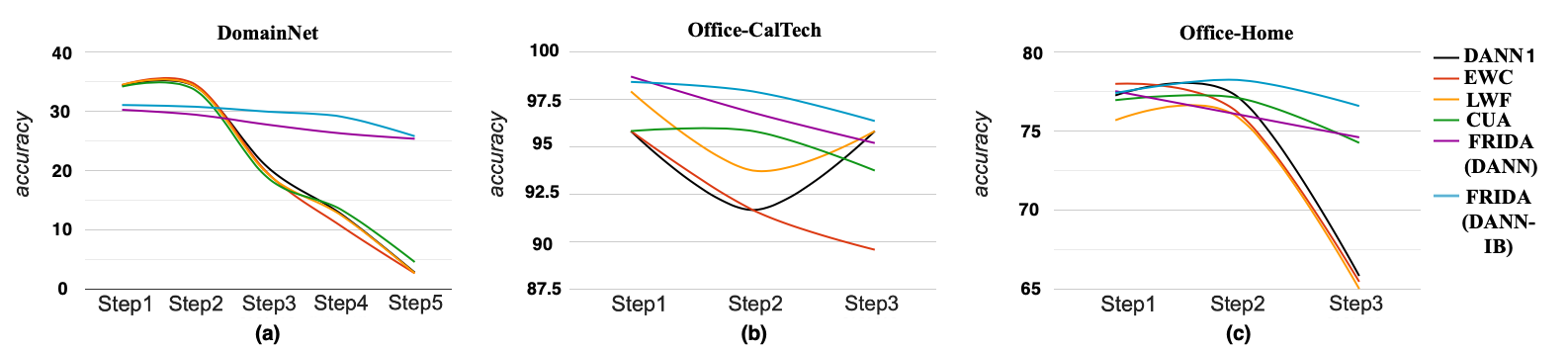}
    \caption{The  plots depicting the evolution of the test accuracy for $\mathcal{T}_1$ for (a) DomainNet (b) Office-CalTech, (c) Office-Home for all the episodes. Drop in accuracies between two consecutive time frames denotes forgetting. The performance at a time may increase as multiple pseudo labeled domains provide  complementary information for better classifying a domain.}
    \label{fig:t}
\end{figure*}

\begin{figure*}[h]
    \centering
    \includegraphics[width=1\linewidth, height=3.7cm]{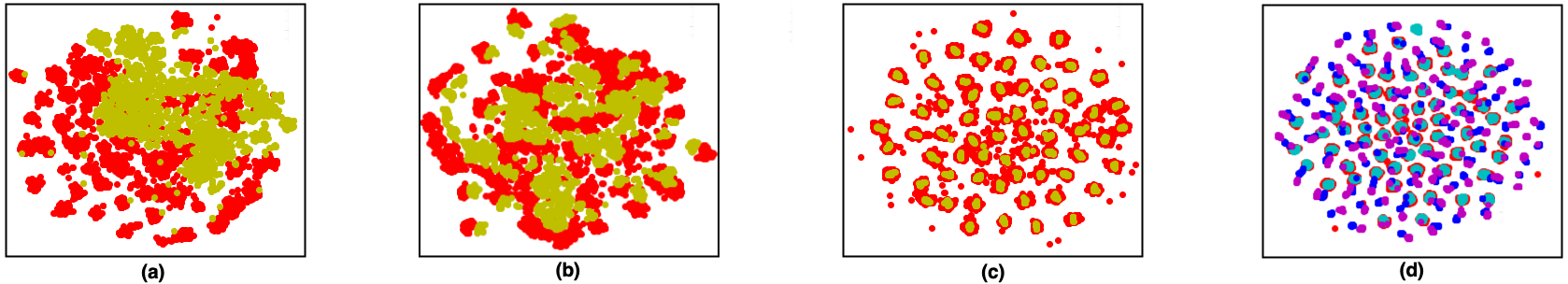}
     \caption{Qualitative analysis of the DGAC-GAN outputs at $\tau=3$ (which is trained on $\tilde{R}^2 \cup \tilde{P}^2 \cup \tilde{C}^2 \cup \hat{A}_3$)
     for Office-Home. We show the t-SNE plots of the real (red) and generated (green) samples for the domain C for (a) the model without both $\mathcal{R}^{gan}$ and domain label in $f_{sh}^{gan}$ (baseline AC-GAN model), (b) the model with the domain label in $f_{sh}^{gan}$ but without $\mathcal{R}^{gan}$, (c) The full DGAC-GAN model. (d) t-SNE plot for the real and all the synthetic samples for domain R (aka $\mathcal{S}$) for Office-Home where the four colors show the samples available / generated at different $\tau$. }
        \label{fig:t1}
\end{figure*}

\begin{figure*}[h]
    \centering
    \includegraphics[width=1\linewidth, height=3.5cm]{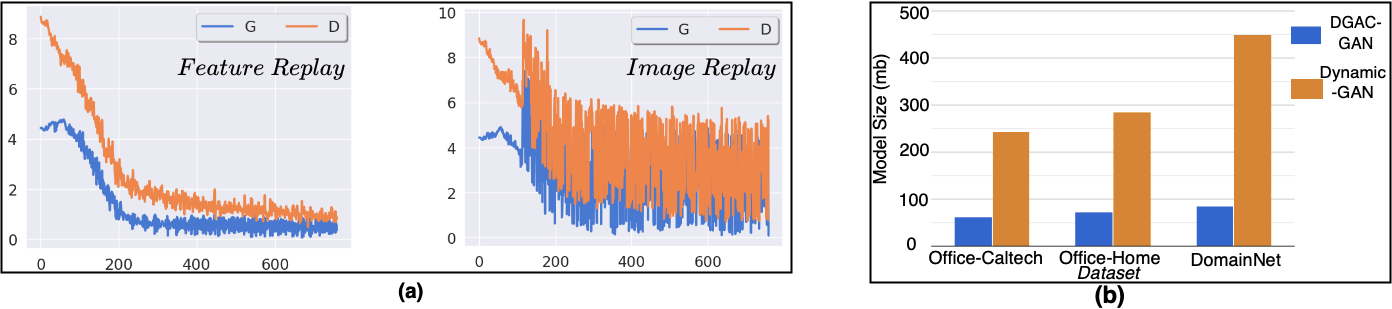}
     \caption{ (a) The convergence comparison of feature and image replay with DGAC-GAN. (b) Disk space comparison of DGAC-GAN and the dynamically expanded incremental GAN.}
        \label{fig:tt1}
\end{figure*}

\section{Experiments}

\noindent \textbf{Datasets}: We evaluate FRIDA on three benchmark DA datasets: Office-Home (\cite{47}), Office-CalTech (\cite{49}), and DomainNet (\cite{48}), respectively. Here we mention the details of the datasets: i) \textit{Office-Home}: The office-home data consists of four visual domains, Art (A), Clipart (C), Real World (R), and Product (P) each consisting of images from $65$ visual categories and a total of $15,500$ images are present over all the domains. ii) \textit{Office-CalTech}: The Office-Caltech is an extension of the Office-31 data where the $10$ shared classes with the CalTech-256 dataset are considered to obtain four domains: Webcam (W), DSLR (D), Amazon (A), and CalTech (C). iii) \textit{DomainNet}: DomainNet is a large-scale dataset with six visual domains consisting of $0.6$ million images from $345$ categories. The domains are: Clipart (C), Real (R), Infograph (I), Painting (P), Sketch (S), and Quickdraw (Q). We randomly choose the source and the sequence of the target domains for all the datasets. The orders are as follows: $R \rightarrow P \rightarrow C \rightarrow A$ (Office-Home), $A \rightarrow D \rightarrow W \rightarrow C$ (Office-CalTech), and $R \rightarrow P \rightarrow C \rightarrow S \rightarrow Q \rightarrow I$ (DomainNet). The first domain works as $\mathcal{S}$ while the rests denote the $\{\mathcal{T}_{\tau}\}_{\tau=1}^T$. For all the domains, we randomly select $30\%$ samples to define the test sets while the remaining data are used during training. We consider the $2048$ dimensional feature representations extracted from the Imagenet pre-trained Resnet-50 (\cite{50}).

\noindent \textbf{Model architecture and training protocol}: In DGAC-GAN, the generator $\mathcal{G}^{gan}$ and the shared part of the discriminator $f_{sh}^{gan}$ are implemented as multi-layer perceptrons (MLP) with three layers each.
The latent variable $\textbf{z}$ (Section 3.2) has a dimensionality of $2000$ in all the experiments whereas the domain identifier $\tau$ is described by three-bit binary encoding. The discriminator classifiers $(f_D^{gan}, f_{C}^{gan})$ are defined in terms of single dense layers. For  DANN-IB, the feature extractor $\mathcal{G}^{dann}$ is represented by a three-layer MLP with latent space dimensionality of $256$ while the classifiers $f_C^{dann}$ and $f_D^{dann}$ have single dense layer each.  For pseudo-labeling in DANN-IB, we fix the confidence threshold $Th$ to be $0.95$ in order to ensure that the selected samples closely approximate the true labels. Besides, we generate $100$ samples per-class for all the previous domains while carrying out the adaptation.
Both the networks are trained using adam (\cite{52}) with a learning rate of $0.001$, $\beta_1=0.5, \beta_2=0.9$, and a batch size of $64$, respectively.

\noindent \textbf{Evaluation protocols}: We report the average performances for all the source and the target domains over all the time stamps defined by $A(\mathcal{T}_{\tau})$ for the domain $\mathcal{T}_{\tau}$. Subsequently, the mean over all the domain-specific accuracies are mentioned for the entire dataset.We report a measure of forgetting to highlight the evolution of the test performances for a given domain over the time. Precisely, for a domain $\mathcal{T}_{\tau}$, let $\mathcal{A}_{\tau}$ and $\mathcal{A}_{\tau+1}$  denote the performances at time $\tau$ and $\tau + 1$.Given that, the average forgetting for $\mathcal{T}_{\tau}$ till time $T$ is defined by $F(\mathcal{T}_{\tau})$ whereas the mean forgetting over all the domains is measured by averaging the $F(\mathcal{T}_{\tau})$ scores over all the domains for a given dataset. The mean forgetting over all domains is given by $ F(\mathcal{T}_{\tau}) = \frac{1}{T-\tau+1} \underset{k=\tau -1}{\overset{T-1}{\sum}} \mathcal{A}_{k+1} - \mathcal{A}_k$. A positive $|\mathcal{A}_{\tau+1} - \mathcal{A}_{\tau}|$ indicates that the test accuracy increases at time $\tau + 1$ than at time $\tau$ for $\mathcal{T}_{\tau}$.


 
\begin{figure*}[h]
    \centering
    \includegraphics[width=1\linewidth, height=4cm]{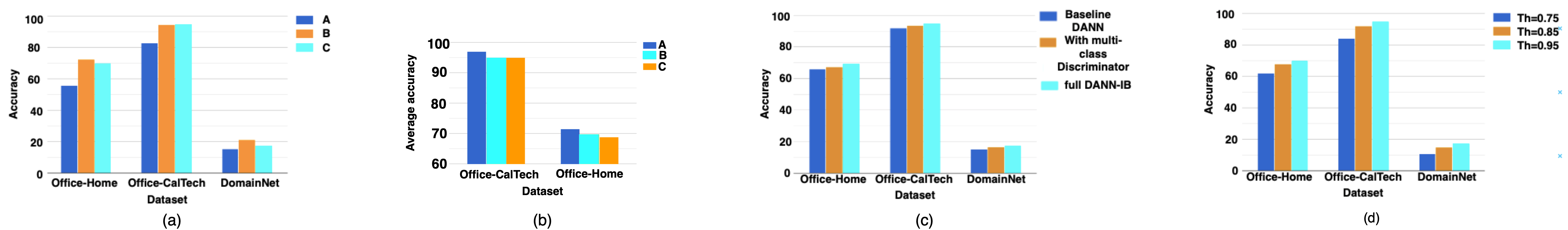}
    \caption{(a) The performance analysis of A - FRIDA (DANN-IB) (without feature replay), B - Proposed non-rehearsal based FRIDA (DANN-IB), C - FRIDA (DANN-IB) with rehearsal based replay where the original pseudo-labeled samples are stored. (b) The sensitivity analysis on the number of generated samples per class- 'A': $200$, 'B':$100$, and 'C':$50$. (c)Ablation study of the models: baseline DANN.  DANN with multi-class discriminator, and DANN-IB for the three datasets. (d) Importance of the threshold $Th$ parameter for pseudo-labeling on the final average accuracy scores.   }
        \label{fig:t3}
\end{figure*}

\begin{figure*}[h]
    \centering
    \includegraphics[width=1\linewidth , height=3.4cm ]{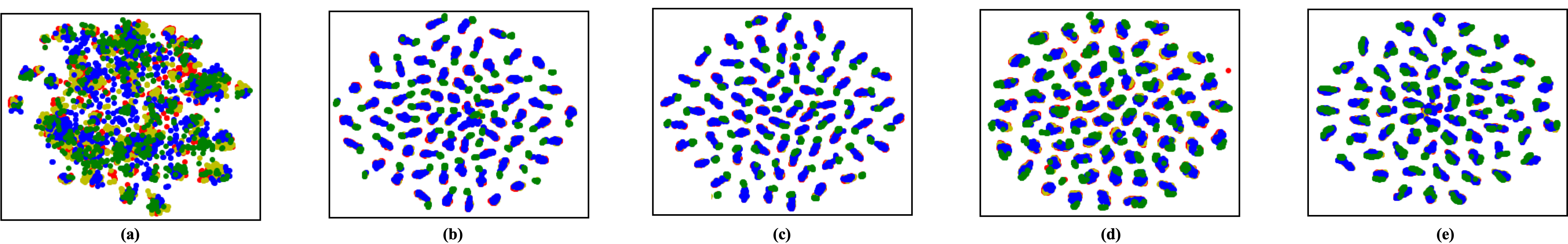}
    \caption{Qualitative performance comparison between DANN and DANN-IB trained to adapt the target domain A given the auxiliary source domain $R \cup P \cup C$ (in different colors). (a)the original resnet-50 features. (b)Features generated by DGAC-GAN. (c)After adaptation by DANN. (d) t-SNE of $\mathcal{C}+1$-class discriminator DANN. (e)After adaptation by DANN-IB.}
        \label{fig:t2}
\end{figure*}


\subsection{Comparison to the literature}

Owing to limited existing works for IDA and the fact that no prior work conforms to the proposed problem setup, we customize the existing DA/IL algorithms for the IDA setup and consider them for comparative analysis. We consider two baseline cases using DANN1 \label{Dann1}: i) we assume $\mathcal{S}$ is present all over and perform adaptation between $\mathcal{S}$ and $\mathcal{T}_{\tau}$ at every $\tau$.  This is referred to as DANN2 \label{Dann2} and denotes the case of IDA without replay. ii) Both $\mathcal{S}$ and the unlabeled data from the past domains are stored and the adaptation goes in this way - $\mathcal{S} - \mathcal{T}_1$, $\mathcal{S} - \mathcal{T}_1 \cup \mathcal{T}_2$ etc. The setup is resembles the notion of multi-target DA and is termed DANN 2. We consider another such technique called DADA (\cite{63}) as well.
Next, we integrate DANN with the elastic weight consolidation (EWC) (\cite{12}) and learning without forgetting (LwF) (\cite{25}) techniques for preventing forgetting of the previous domains for incrementally training DANN. While we introduce the EWC regularizer between $DA_{\tau}$ and $DA_{\tau-1}$ to control the deviation of the important parameters, a selected set of replay samples are considered from the previous domains to define the distillation loss for LwF. Besides, we consider two existing IDA approaches called CUA (\cite{11}) and IADA (\cite{44}) both of which use the source domain information and feature replay together at every step, hence, they are different from our setup. Both of them consider DANN as the DA model. Also, IADA projects the target domains into the distribution of the source domain while CUA learns a shared feature space for both the domains. 
We report the values of FRIDA using both DANN-IB and DANN. As can be observed in Table \ref{tab:1}-\ref{tab:3}, FRIDA (both with DANN and DANN-IB) is able to consistently outperform all the other methods by a substantial margin. In particular, FRIDA (DANN) provides a value of $15.61 \%$ for DomainNet, $92.80 \%$ for Office-CalTech, and $68.55 \%$ for Office-Home for the target domains which are better than all the competing techniques. We note that DANN 2 considers the setup where the target domains are merged and this setup is expected to provide high performance. Nonetheless, such a setup is unrealistic in IDA as the domain information becomes non-existent after the respective time stamp. Also, the performances of DANN 2 and IADA for the source domain are always fixed and high as no forgetting is incurred.
For techniques like  EwC, LwF or CUA, we believe that the respective  regularization techniques prevent the models to diverge much with respect to the previous domains. However, such a restriction hinders the adaptation process for the new domains as the domain distributions can vary significantly in IDA. To resolve this issue, FRIDA proposes to re-train the DA module with a combination of the synthetic past data and present raw data. We find FRIDA (DANN-IB) beats FRIDA (DANN) consistently by producing average accuracies of $17.34 \%$ for DomainNet, $95.05 \%$ for Office-CalTech, and $69.82 \%$ for Office-Home, respectively. The performance for the source domains are also enhanced with FRIDA(DANN-IB) by $\geq 20\%$ for DomainNet and $\geq 5 \%$ for Office-CalTech with respect to DANN1 / CUA  / EWC / LwF. Figure \ref{fig:t} depicts the changes in the performance for the domain $\mathcal{T}_1$ for all the datasets(Table: \ref{tab:14} -\ref{tab:34}at Appendix.1,  shows the performance of all the doamins for each datasets   ). The evolution for the other domains are mentioned in the supplementary. It is seen that FRIDA maintains the best stability-plasticity amongst all the approaches where the EWC, CUA, LwF show sharp sudden degradation. 

\subsection{Critical analysis}

\begin{figure*}[h]
    \centering
    \includegraphics[scale=.11]{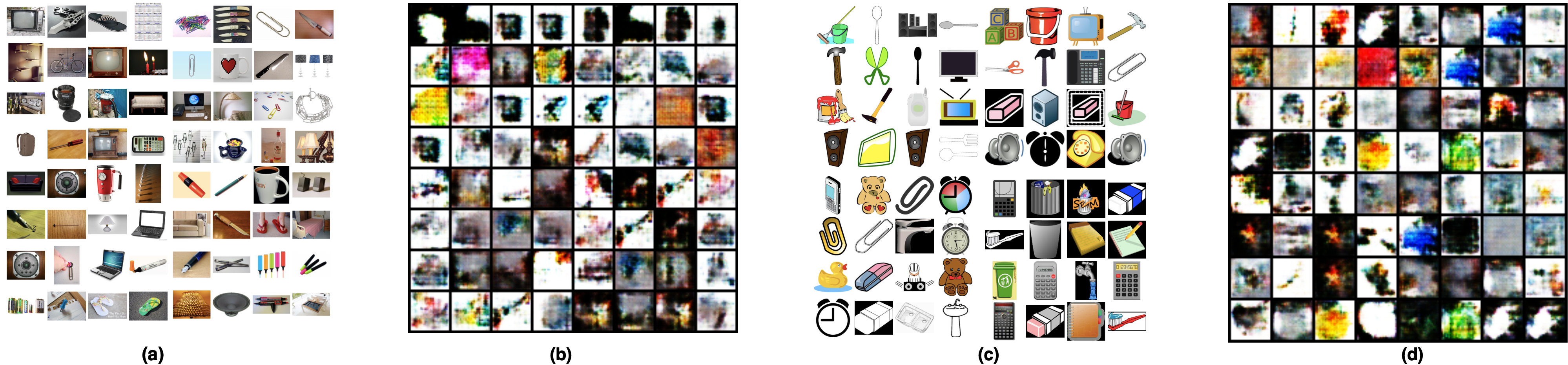}
    \caption{The results of image generation using DGAC-GAN for the (a-b) Real-World domain and where (a) shows the original images and (b) shows the ones generated by DGAC-GAN, (c-d) ClipArt domain where (c) shows the original images and (d) shows the images generated by DGAC-GAN, both for the Office-Home dataset.}
    \label{fig:image_gen}
\end{figure*}

\noindent \textbf{Analysis of DGAC-GAN}: We showcase a qualitative study on the efficacy of the DGAC-GAN model in generating discriminative samples when DGAC-GAN is trained at $\tau=3$ for Office-Home. We consider three model variants for DGAC-GAN and train three experimental scenarios here as follows: i) without conditioning the discriminator with $\tau$ and without $\mathcal{R}^{gan}$, ii) without $\mathcal{R}^{gan}$ only, and iii) the full DGAC-GAN model, respectively. Figure \ref{fig:t1}(a-c) depict the t-SNE plots for the Clipart domain for all the three models. It can be understood that both the domain identifier and $\mathcal{R}^{gan}$ together direct the generation of discriminative synthesized samples. Besides, Figure \ref{fig:t1}(d) illustrates the t-SNE combining the real and the synthesized samples generated at $\tau=1-3$ for the domain R (used to define $\mathcal{S}$) of Office-Home. Here we would like to emphasize that the samples generated at different stages are classwise overlapped with the real samples. This confirms that the joint training strategy followed in DGAC-GAN controls forgetting. As we mentioned, the feature replay in DGAC-GAN offers a stable training than image replay (Figure \ref{fig:tt1}(a)). Further, we compare DGAC-GAN with a dynamically expanded GAN model where new domain-specific generators and discriminators are added. As can be observed from Figure \ref{fig:tt1}(b), DGAC-GAN is extremely lightweight but produces equivalent performance with the dynamically expanded GAN. (More qualitative results are in appendix.4)

It is observed that the image generation of multi domain is not full-field the requirement. We have generated the images of Office-Home, Figure \ref{fig:image_gen} and it can be seen that the generated images are quite bad and failed to mimic the original images. So we have done feature replay instead of image replay. 

\noindent \textbf{Role of replay memory}: Moreover, we assess the need of memory replay in IDA and perform an experiment without introducing the notion of replay in FRIDA which mimics the setup of DANN 1 but we use DANN-IB instead. We also consider the case where the original pseudo-labeled samples are saved at every time instead of training the DGAC-GAN. We find from Figure \ref{fig:t3}(a) that while FRIDA beats the model without replay considerably by at least $5\%$ for all the data,  the performance is at par or decreases marginally by $2-3 \%$ when real samples are stored. We remind that in FRIDA, we generate $100$ samples per class using DGAC-GAN while a large number of samples are generated through pseudo-labeling ($>10$K for the data). Figure \ref{fig:t3}(b) shows the sensitivity to the number of generated samples by DGAC-GAN.

\noindent \textbf{Analysis of DANN-IB}: We further analyze DANN-IB model where we train all the four domains of Office-Home together at $\tau=3$. In Figure \ref{fig:t2}(a-b), the t-SNE plots show the distributions of the original resnet-50 features and the synthetic features obtained from DGAC-GAN. Figure \ref{fig:t2}(c-e) depict the visualizations after applying DANN, DANN with multi-class discriminator, and the full DANN-IB on the synthetic samples. We see that DANN-IB is able to produce a more compact classwise alignment among the domains than DANN. We quantitatively evaluate DANN-IB in Figure \ref{fig:t3}(c) where we report the overall average accuracy of FRIDA for three models: baseline DANN, DANN with multi-class discriminator only, and the DANN-IB model, respectively. The inclusions of the multi-class discriminator and $\mathcal{R}_{IB}$ consistently improves the performance of DANN by $1 \%$ each for Office-CalTech and DomainNet. Figure \ref{fig:t3}(d) shows the sensitivity analysis of the method on different threshold values used for the pseudo-labeling stage. It is apparent that a high threshold yields confident predictions (More analysis of DANN-IB is in appendix.5)


\section{Conclusions}

We tackle the data-free IDA problem and propose a novel generative feature replay based framework called FRIDA. The working of FRIDA alternates between synthetic data generation for old domains using DGAC-GAN and alignment of the old and new domains using DANN-IB. Both the DGAC-GAN and DANN-IB models can be updated incrementally as new domains arrive. Experimental results on three benchmark datasets confirm the efficacy of FRIDA. Currently, FRIDA assumes a closed-set scenario where all the domains are assumed to contain the same set of classes. We are interested in extending it to support open-set scenario where the new domains may contain novel classes.


\bibliographystyle{unsrt}  
\bibliography{references}  
\appendix
\section*{Appendix}

\subsection{Per domain accuracy for all the datasets at different timestamps}\label{appen1}

Here we show the evolution of the test performances for different source and target domains for all the datasets considered. Table \ref{tab:14} - \ref{tab:34} depict the same for Office-Home, Office-CalTech and DomainNet respectively. The order of the domains for Office-Home, Office-CalTech, and DomainNet are mentioned in the main paper.

\begin{table}[h]
\centering
\begin{tabular}{|c|c|c|c|}
\hline
& Step 1 & Step2 & Step 3\\
\hline
$\mathcal{S}$ &84.02 &83.03  & 82.42 \\
\hline
$\mathcal{T}_1$ &77.4 & 78.23 & 76.58 \\
\hline
$\mathcal{T}_2 $ &- &63.28  & 65.34 \\
\hline
$\mathcal{T}_3$ &- & - & 67.76 \\
\hline
\end{tabular}
\caption{Detailed test accuracies at different time instants for the Office-Home dataset.}\label{tab:14}
\end{table}

\begin{table}[h]
\centering
\begin{tabular}{|c|c|c|c|}
\hline
& Step 1 & Step2 & Step 3\\
\hline
$\mathcal{S}$ &94.75 &95.13  &95.21  \\
\hline
$\mathcal{T}_1$ &98.42 & 98.23 & 96.36 \\
\hline
$\mathcal{T}_2 $ &- & 100 & 98.13 \\
\hline
$\mathcal{T}_3$ & -& - & 88.42 \\
\hline
\end{tabular}
\caption{Detailed test accuracies at different time instants for the Office-CalTech dataset.}\label{tab:24}
\end{table}


\begin{table}[ht]
\centering
\begin{tabular}{|c|c|c|c|c|c|}
\hline
& Step 1 & Step2 & Step 3 & Step4 & Step5\\
\hline
$\mathcal{S}$ &68.31 & 67.41 & 64.71& 64.35 & 58.79 \\
\hline
$\mathcal{T}_1$ &31.01 &30.76  &29.92 & 29.08 & 26.35 \\
\hline
$\mathcal{T}_2 $ & -&29.45  &28.30  & 28.71 & 26.30\\
\hline
$\mathcal{T}_3$ &- & - & 17.50&17.19  & 15.68\\
\hline
$\mathcal{T}_4$ &- &-  &- & 2.74 & 2.94\\
\hline
$\mathcal{T}_5$ &- & - & -& - & 8.98\\
\hline
\end{tabular}
\caption{Detailed test accuracies at different time instants for the DomainNet dataset.}\label{tab:34}
\end{table}

\subsection{Convergence of MADA, DANN and DANN-IB on IDA (Office-Home)}

\begin{figure}
    \centering
    \includegraphics[scale=.6]{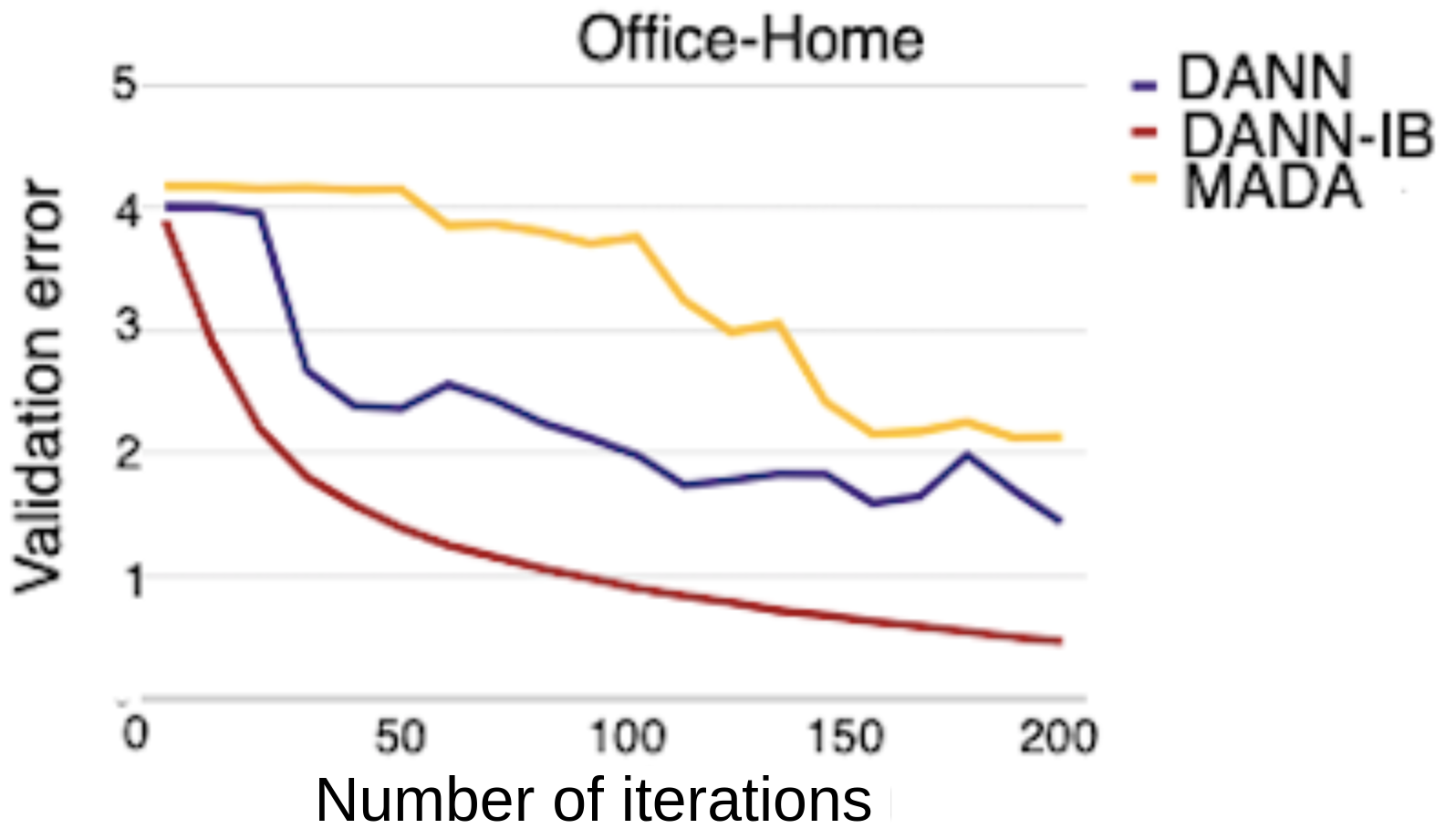}
    \caption{Convergence of MADA and DANN-IB on IDA (Office-Home).}
    \label{fig:22}
\end{figure}

Unlike DANN's single discriminator with a binary output, DANN-IB uses a single discriminator having $\mathcal{C}+1$ class indices where the $\mathcal{C}+1$th index denotes the initial label for the target domain and MADA uses $\mathcal{C}$ separate binary discriminators one for each of the classes. Training of MADA is driven by the quality of pseudo-labeling which may trigger misclassification and poor convergence in IDA. DANN-IB training does not use pseudo-labels. To the best of our knowledge, DANN-IB is one of the first UDA methods to use the idea of information bottleneck (certainly the first for IDA) which is absent in MADA. Figure \ref{fig:22} shows the convergence of DANN-IB is more smooth and quick with the comparison of MADA and traditional-DANN (\cite{20}).

\subsection{Convergence of losses w/o previous model initialization of FRIDA}

\begin{figure*}[h]
    \centering
    \includegraphics[width=0.95\linewidth, height=3.5cm]{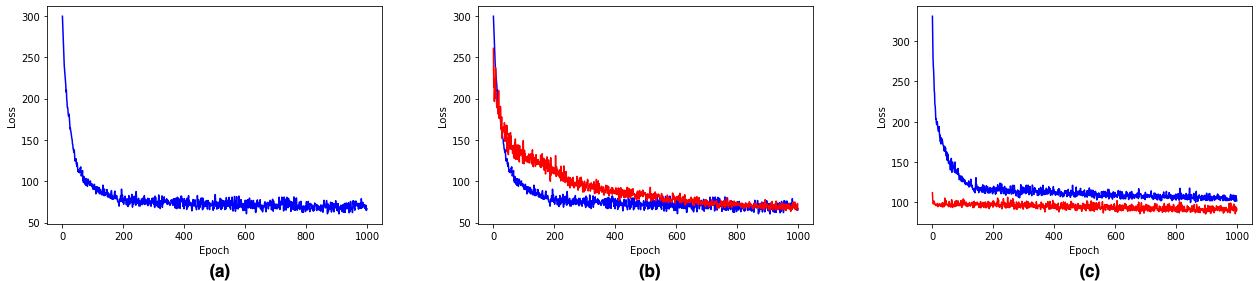}
    \caption{Generator loss of DGAC-GAN for the Office-CalTech dataset at the different timestamp. 'Red' indicates the loss values where DGAC-GAN model is initialized by the previous timestamp model and 'bule' depict the loss when DGAN-GAN is trained by random weight initialization. (a)   Generator loss of $DGAC-GAN_{0}$, with random weight initialization (b) Generator loss of $DGAC-GAN_{1}$, with $DGAC-GAN_{0}$ weight initialization(Red) and random weight initialization(Blue) (c) Generator loss of $DGAC-GAN_{2}$, with $DGAC-GAN_{1}$ weight initialization(Red) and random weight initialization(Blue)} 
        \label{fig:gan_generator_loss}
\end{figure*}

\begin{figure*}[h]
    \centering
    \includegraphics[width=0.95\linewidth, height=3.5cm]{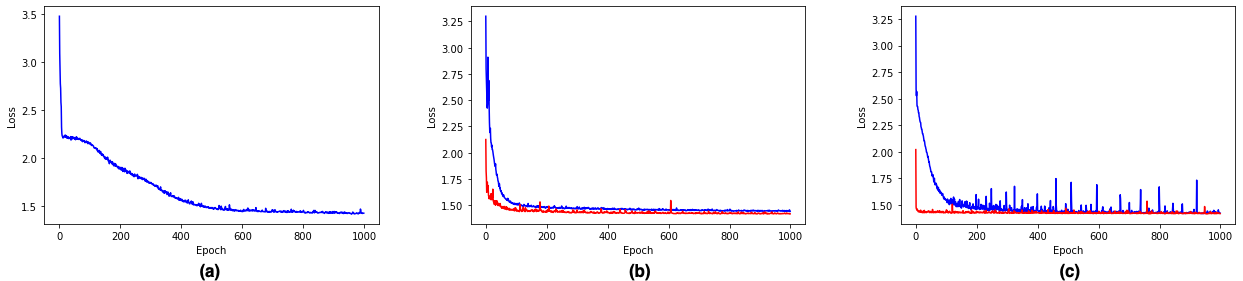}

\caption{ Discriminator loss of DGAC-GAN for the Office-CalTech dataset at the different timestamp. 'Red' indicates the loss values where DGAC-GAN model is initialized by the previous timestamp model and 'bule'  depict the loss when DGAN-GAN is trained by random weight initialization. (a)Discriminator loss of $DGAC-GAN_{0}$, random weight initialization(Blue) (b) Discriminator loss of $DGAC-GAN_{1}$, with $DGAC-GAN_{0}$ weight initialization(Red) and random weight initialization(Blue) (c) Discriminator loss of $DGAC-GAN_{2}$, with $DGAC-GAN_{1}$ weight initialization(Red) and random weight initialization(Blue)} 
        \label{fig:gan_discriminator_loss}
\end{figure*}

\begin{figure*}[h]
    \centering
    \includegraphics[width=0.95\linewidth, height=3.5cm]{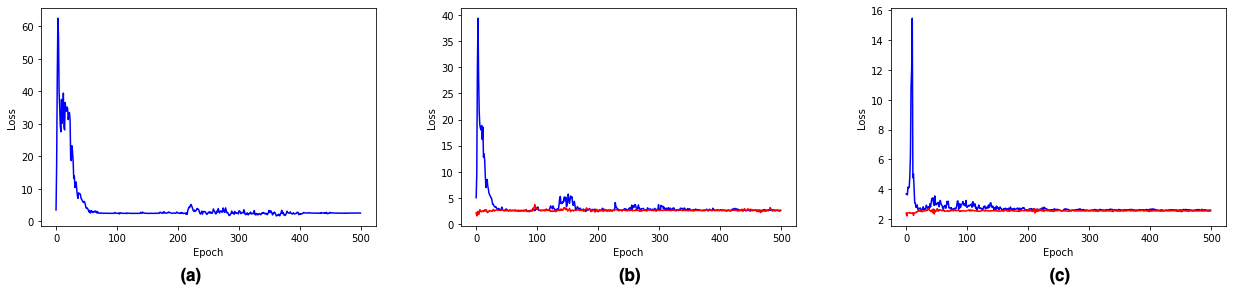}
    
    \caption{ Total loss of $DANN-IB$ for the Office-CalTech dataset at the different timestamp. 'Red' indicates the loss values where the model is initialized by the previous timestamp DANN-IB model and 'bule' depicts the loss when $DANN-IB$ is trained by random weight initialization. (a)  Loss of $DANN-IB$ $(DA_1)$(main paper), with random weight initialization (b) Loss of $DANN-IB$ $(DA_2)$(main paper), with $(DA_1)$(main paper) weight initialization (Red) and random weight initialization(Blue) (c) Generator loss of $DGAC-GAN$ $(DA_3)$(main paper), with $(DA_2)$(main paper) weight initialization(Red) and random weight initialization(Blue)}
        \label{fig:dannIB_loss}
 
\end{figure*}

In Figure \ref{fig:gan_generator_loss} and \ref{fig:gan_discriminator_loss} shows the generator and discriminator loss convergence respectively, with and without previous model initialization for GAN generation at different time stamps. The $GAN_{\tau}$ model loss behaves more stable and converges early when $GAN_{\tau}$  model weights are initialized by the $GAN_{\tau-1}$ model weights, $\tau$ is the current timestamp. Figure \ref{fig:dannIB_loss} shows the loss convergence for DANN-IB at different timestamp. We can spot through the Figure \ref{fig:dannIB_loss}, a very significant impact of previous model initialization for loss stability and early convergence of DANN-IB loss, over the DANN-IB model training with random weight initialization.





\begin{figure*}[h]
    \centering
    \includegraphics[width=0.95\linewidth, height=3.7cm]{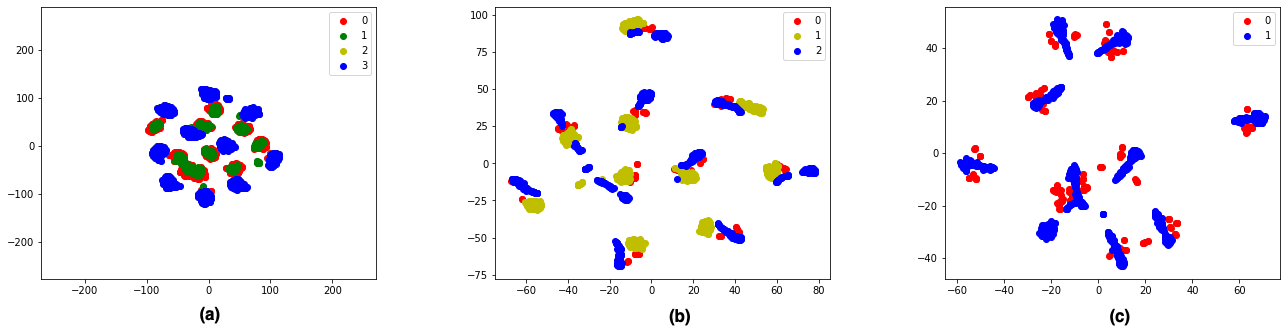}
    \caption{  Available and generated synthetic samples for the Office-CalTech dataset (a) samples for $\mathcal{S}$ where red color denotes the original samples and the other colors correspond to the synthetic samples generated at different times, (b) Original and generated samples for domain $\mathcal{T}_1$, (c) Original and generated samples for domain $\mathcal{T}_2$.} 
        \label{fig:all1}

\end{figure*}


\subsection{More qualitative results on DGAC-GAN}

In Figure \ref{fig:all1}(a-c), we show the t-SNE plots for the original samples and the synthetic samples generated at different time stamps for $\mathcal{S}$, $\mathcal{T}_1$, and $\mathcal{T}_2$ for Office-CalTech. It can be observed that the samples are class-wise overlapped.  In Figure \ref{fig:all1}(a-c), the colors (green, yellow and blue) represents the generated samples between $1 \leq \tau \leq 3$ and red denotes the original samples in all the cases but the index (0-3) represents the number of times sample generated throughout the incremental steps. For example, in figure \ref{fig:all1}(a-c), blue represents $\tau=3$ and the samples of $\mathcal{S}$ has been generated three times till that step but for $\mathcal{T}_1$, and $\mathcal{T}_2$ it has been generated twice and once, respectively.

\subsection{Theoretical analysis for the DANN-IB model}

Here we show that the inclusion of the multi-class discriminator and the information bottleneck regularizer helps in obtaining a tighter bound for the target risk in the standard adversarial unsupervised domain adaptation framework like DANN. Given the hypothesis instances $h, h' \in \mathcal{H}$ where $\mathcal{H}$ defines the hypothesis space for the source domain features $\mathcal{S}(X_s)$ and the target domain data $\mathcal{T}(X_t)$, the test error ($\epsilon_t$)  for $\mathcal{T}$ is upper-bounded as follows \cite{5},

\begin{equation}
    \epsilon_t \leq d_{\mathcal{H}\Delta \mathcal{H}}(X_s, X_t) + \epsilon_s(h) + \underset{h' \in \mathcal{H}}{\text{min}} \epsilon_t(h') + \epsilon_s(h')  
\end{equation}

\noindent where $\epsilon_s$ is the test error function for the source domain. $d_{\mathcal{H}\Delta \mathcal{H}}$ denotes a measure of discrepancy between the domains and is expressed as,

\begin{equation}
    d_{\mathcal{H}\Delta \mathcal{H}} = 2 \underset{h,h' \in \mathcal{H}}{\text{sup}} ||\mathbb{E}_{x \in X_s}[h(x) \neq h'(x)] - \mathbb{E}_{x \in X_t}[h(x) \neq h'(x)]||
\end{equation}

As per the information bottleneck principle (\cite{43}), we aim to minimize the mutual information $I$ between the raw input $X$ with label $Y$ and the latent representation $Z$: $I(X,Z)$ so that $I(Z,Y)$ is simultaneously maximized. As a result, only the task-specific features are retained in $Z$ ignoring the domain specific artifacts from $X$. \cite{43} further shows that the difference between the training and test error is bounded by a monotonic function of $I(X,Z)$. Naturally, the bottleneck principle aids in minimizing $\epsilon_s(h)$ by ensuring that only the feature embeddings consistent with the label information are to be retained. 
This, in turn, imposes a tighter upper-bound to $\epsilon_t$ since $\epsilon_s$ is properly minimized. Furthermore, the use of $\mathcal{R}_{IB}$ (Eq. \ref{eqn:four}) constraints that the latent distributions for the source and the target domains should be consistent with the standard normal distribution as per the variational upper bound principle: $d_{\mathcal{H}} (\mathcal{S},\mathcal{T}) \leq d_{\mathcal{H}} (\mathcal{T}, \mathbb{N}(0,\mathbb{I})) + d_{\mathcal{H}} (\mathcal{S}, \mathbb{N}(0,\mathbb{I}))$. We note that this also satisfies the notion of triangular inequality for the $d$-divergence. Hence, minimizing $\mathcal{R}_{IB}$ in Eq.(\ref{eqn:five})  alternately minimizes $d_{\mathcal{H} \Delta \mathcal{H}}$. As a result, we obtain a stricter upper bound for $\epsilon_t$.

On the other hand, the $\mathcal{C}+1$-class discriminator $f_D^{dann}$ helps in better aligning both the domains since it tries to assign the target samples in one of the modes of the source data distribution. This is as opposed to the DANN theory which performs global distributions matching between both the domains using a binary classifier. Ideally, it is possible to have a latent space in DANN where both the domains lie on the same side of the hyperplane but the samples are not overlapped due to a trivial mapping. The proposed multi-class discriminator solves such a scenario. Further, any misclassification for the source domain samples is avoided in this case given the classification loss $\mathcal{L}_c$ Eq.(\ref{eqn:five}). In gist, DANN-IB proposes a better bound for the target domain samples, which is also empirically reflected in Table \ref{tab:1}-\ref{tab:3}.

\end{document}